\documentclass[conference]{IEEEtran}
\IEEEoverridecommandlockouts

\usepackage{cite}
\usepackage{amsmath,amssymb,amsfonts}
\usepackage{algorithmic}
\usepackage{graphicx}
\usepackage{textcomp}
\usepackage{xcolor}

\usepackage{comment}
\usepackage{caption}
\usepackage{subcaption}
\usepackage{multirow}
\usepackage{balance}
\usepackage{hyperref}
\usepackage{url}
\usepackage{booktabs}
\usepackage[export]{adjustbox}
\usepackage{tabulary}
\usepackage{soul}
\usepackage{flushend}
\usepackage{color}
\usepackage{pifont}

\def\BibTeX{{\rm B\kern-.05em{\sc i\kern-.025em b}\kern-.08em
    T\kern-.1667em\lower.7ex\hbox{E}\kern-.125emX}}
    
\begin{document}

\title{DVM-CAR: A Large-Scale Automotive Dataset for Visual Marketing Research and Applications}

\makeatletter
\newcommand{\linebreakand}{%
  \end{@IEEEauthorhalign}
  \hfill\mbox{}\par
  \mbox{}\hfill\begin{@IEEEauthorhalign}
}
\makeatother

\author{
\IEEEauthorblockN{Jingmin Huang}
\IEEEauthorblockA{\textit{School of Computing Science} \\
\textit{University of Glasgow}\\
Glasgow, UK \\
2421107h@student.gla.ac.uk}
\and
\IEEEauthorblockN{Bowei Chen}
\IEEEauthorblockA{\textit{Adam Smith Business School} \\
\textit{University of Glasgow}\\
Glasgow, UK \\
bowei.chen@glasgow.ac.uk}
\and
\IEEEauthorblockN{Lan Luo}
\IEEEauthorblockA{\textit{Marshall School of Business} \\
\textit{University of Southern California}\\
Los Angeles, USA \\
lluo@marshall.usc.edu}
\linebreakand
\IEEEauthorblockN{Shigang Yue}
\IEEEauthorblockA{\textit{School of Computer Science} \\
\textit{ University of Lincoln}\\
Lincoln, UK \\
syue@lincoln.ac.uk}
\and
\IEEEauthorblockN{Iadh Ounis}
\IEEEauthorblockA{\textit{School of Computing Science} \\
\textit{University of Glasgow}\\
Glasgow, UK \\
iadh.ounis@glasgow.ac.uk}
}

\IEEEoverridecommandlockouts
\IEEEpubid{\makebox[\columnwidth]{978-1-6654-8045-1/22/\$31.00 ©2022 IEEE \hfill}
\hspace{\columnsep}\makebox[\columnwidth]{ }}
\maketitle
\IEEEpubidadjcol

\begin{abstract}
There is a growing interest in product aesthetics analytics and design. However, the lack of available large-scale data that covers various variables and information is one of the biggest challenges faced by analysts and researchers. In this paper, we present our multidisciplinary initiative of developing a comprehensive automotive dataset from different online sources and formats. Specifically, the created dataset contains 1.4 million images from 899 car models and their corresponding model specifications and sales information over more than ten years in the UK market. Our work makes significant contributions to: (i) research and applications in the automotive industry; (ii) big data creation and sharing; (iii) database design; and (iv) data fusion. Apart from our motivation, technical details and data structure, we further present three simple examples to demonstrate how our data can be used in business research and applications.
\end{abstract}


\section{Introduction}
\label{sec:intro}

The automotive industry is a major industrial and economic force worldwide. There is a rising interest in applying data mining, machine learning and artificial intelligence to automotive research and applications. Marketing researchers team up with automotive designers and computer scientists to solve the research questions and challenges in automotive exterior design, consumer analytics and sales forecasting. These are the key areas of marketing, and solving these research questions and challenges can have positive social consequences for automotive manufacturers and business researchers. On the other hand, data mining, machine learning and artificial intelligence are built upon data, in many cases, massive data. However, business details such as product sales are consistently reserved by companies and are rarely shared with researchers. Besides, data preparation can often be a lengthy, difficult and expensive process for many researchers, particularly, social science researchers' primary focuses are not on programming and database management. Sharing reusable data with the academic community can be an effective and alternative way that makes this process more approachable. 

We present our multidisciplinary initiative of creating a publicly available automotive dataset, which can be used to facilitate business research and market forecasting applications in the automotive market. Our initiative is motivated by the growing interest among the business school in big data and computer scientists are becoming interested in product aesthetics analytics and design but publicly available datasets that cover a wide range of product information, such as product appearance, model specifications and sales data, are extensively lacking. Therefore, we propose a data approach that integrates different online product information to obtain a comprehensive car dataset and hopes our work could boost the emergence of more product datasets for research purpose. In addition, three research samples are provided, illustrating the resulting dataset can comprehensively support the data needs in business analysis, product design and market forecasting studies. It is hoped the shared dataset can be used by researchers and practitioners from different disciplines to conduct economic or business related automotive research and applications. 

This study makes four major contributions. First, in the context of \textbf{domain application}, our dataset meets the growing need for a comprehensive automotive dataset for economic and business research. For example, our dataset can be used for automotive exterior design and consumer analytics, which will benefit automotive manufacturers, help them better predict their targeted consumer segment preferences, and then use the obtained insights to direct the design of new exterior stylings. Our dataset can also be used for car sales forecasting which benefits all the participants in the automotive ecosystem including car dealers, consumers and marketers. Second, from a perspective of \textbf{big data}, to the best of our knowledge, our developed dataset is {the} very first large-scale automotive dataset. It contains 1.4 million images from 899 different car models and corresponding specification and sales information from over ten years in the UK market. The four characteristics of big data (i.e., volume, variety, velocity, and veracity) are satisfied, which makes the dataset can be used for different types of analytics and forecasting tasks. The dataset can be used by multidisciplinary researchers to solve different tasks. Many interesting insights, data-driven models and forecasting applications can be derived. Third, in terms of \textbf{database design}, we conduct a survey study with the researchers working in either business or computer science field. The data challenges these researchers meet can be categorized into three major issues: coverage, accessibility and quality. We design the proposed dataset by addressing these issues with the hope that our data can be researcher-friendly and thus can have a large impact in research and practice. We also demonstrate a good practice of developing a dataset which can alleviate the common data issues faced by researchers. Last but not least, our work contributes to the multi-source \textbf{data fusion} as it includes different data formats from different sources. Car images, model specification and sales information are collected from different online sources while they are merged and stored in a flexible and hierarchical structure that allows it to be easily expanded with new data and used by researchers. 


\section{Data Collection and Cleaning}
\label{sec:data_preparation}
Nowadays researchers often collect online data and process them as new datasets for their research. Datasets like ImageNet~\cite{deng2009imagenet} and OpenImages~\cite{Kuznetsova2020} are characterized by their enormous size and have achieved massive success in computer science research and applications. The two existing car image datasets, Stanford-Car~\footnote{\href{https://ai.stanford.edu/~jkrause/cars/car\_dataset.html}{https://ai.stanford.edu/\~{}jkrause/cars/car\_dataset.html}}\cite{krause20133d} and CompCars~\footnote{\href{http://mmlab.ie.cuhk.edu.hk/datasets/comp\_cars/index.html}{http://mmlab.ie.cuhk.edu.hk/datasets/comp\_cars/index.html}}\cite{yang2015large}, are also based on the web scraped contents. Motivated by these existing works, we develop the dataset by collecting and integrating data from different online sources. First, car images are collected from the popular automotive classified advertising platforms. These platforms are popular online marketplaces in the UK for buying and selling used cars, which host millions of car images and their selling prices for almost all car models from different automotive manufacturers. Second, car sales data is collected from the Driver and Vehicle Licensing Agency (DVLA), which is part of the Department for Transport, holding over 49 million driver records and over 40 million vehicle records in the UK. The DVLA publishes the statistics of newly registered vehicles in its seasonal reports. We extract and add the sales appropriately for various car models. Third, new car prices are collected from car review websites, which cover the selling prices of various car trims sold in past years. 

In the preparation of data sharing, a number of data cleaning and fusion steps are performed. First, to comply with the General Data Protection Regulation (GDPR)~\cite{VoigtPaul2017}, we use machine learning techniques to detect and remove privacy or unnecessary information. Mask R-CNN~\cite{he2017mask} (trained with COCO~\cite{Lin2014}) is used to detect the appearing objects in car images. It helps to locate the positions of car bodies and detect whether other objects such as human body appear in raw images. Images with unexpected contents are dropped, and remaining images are cropped according to detected car positions. These image pre-processing steps aim to ensure resulting images are primarily occupied by car bodies without other unnecessary information. Moreover, we apply the algorithm from~\cite{silva2018a} to detect registration plates and then blur corresponding regions. Second, machine learning models are used to filter out non-exterior-viewing images. An ImageNet pre-trained convolutional neural network (CNN)~\cite{lecun1998gradient} is fine-tuned with manually prepared samples where pictures are labelled according to their qualities and observation viewpoints. The trained CNN classifies images according to observation viewpoints. As illustrated in Fig.~\ref{fig:three_graphs}, only images taken from the eight targeting viewpoints are kept for later usage. This reduces the size of the raw image set from over six million to less than 1.5 million. Third, all the non-visual contents are processed into attributes and stored as data tables. The ambiguous values of observations are unified or corrected. All car models are assigned unique identifiers (IDs) used for information integration. Information such as annual sales is aggregated into more abstract levels. Images are listed in the table with the observation viewpoints and source advert IDs.

\section{Towards a Good Research Dataset}
\label{sec:Survey}

During the data preparation, we deploy a survey using Qualtrics to explore the most common data issues that people face in their research and practice. We collect responses from 54 researchers, including 26 participants with a computer science background and 28 with a business studies background, including economics, marketing and management. The reported issues from our survey study can be broadly divided into three categories: coverage, accessibility, and quality. 

\textbf{Coverage} refers to the issue of a dataset not containing the needed information. This happens in two situations. First, the dataset is not comprehensive enough. It does not contain the attributes or variables that a researcher is interested in for a specific study. Second, the dataset does not cover enough records or samples. Our dataset is designed to have excellent coverage of both variables and observed samples. For the former, as presented in Table~\ref{tab:target_variables}, we design the dataset to contain many important car specifications and sales variables used in the related marketing studies. In addition to these variables, as also demonstrated in Fig.~\ref{fig:three_graphs}, we provide corresponding images for various car models in various angles over a long-term period. The existing public car image datasets, such as the Stanford-car and CompCars datasets, do not include the mentioned car specification and sales variables as they are mainly designed for computer vision tasks. In terms of volume, the dataset covers millions of registered cars in the UK market in the past decades.

\begin{table}[t]
\small
\centering
\caption{Summary of car specification and sales variables in related studies and our dataset.}
\label{tab:target_variables}
\begin{tabular}{p{1.08in}|c|c|c|c|c}
\toprule
\multirow{2}{*}{Variables/features} & \multirow{2}{*}{\begin{tabular}[c]{@{}c@{}}Our  \\ dataset\end{tabular}} & \multicolumn{4}{c}{Related work} \\
\cline{3-6}
    &  &  ~\cite{kukova2016we} & ~\cite{korenok2010non} & ~\cite{jindal2016designed} & ~\cite{landwehr2011} \\
\toprule
Sales/market share & \checkmark  & &  \checkmark & \checkmark & \checkmark \\
Price & \checkmark & \checkmark & \checkmark & \checkmark & \checkmark \\
Brand & \checkmark & \checkmark & \checkmark & & \checkmark \\
Exterior features & \checkmark & \checkmark & \checkmark & \checkmark & \checkmark \\
Fuel economy & \checkmark & \checkmark & & \checkmark & \\
Horsepower  & \checkmark & \checkmark & & \checkmark & \\
Engine & \checkmark & \checkmark & & \checkmark & \\
Transmission & \checkmark & \checkmark & & \checkmark & \\
Equipment features & \checkmark & \checkmark & & & \checkmark\\
Life-cycle/model year & \checkmark & & \checkmark & \checkmark & \checkmark \\
Advertising & & & \checkmark & \checkmark & \\
Reliability & & \checkmark & \checkmark & & \\
Safety & & \checkmark & \checkmark & & \\
Driving/handling & & \checkmark & & \checkmark & \\
Ergonomics/rooming & & \checkmark & & \checkmark & \\
Interior & & \checkmark & & & \\
\bottomrule
\end{tabular}
\end{table}

\textbf{Accessibility} refers to the difficulties regarding data usage caused by {closed or proprietary datasets} or {the underlying complex data structure}. Our participants in the survey {highlighted} that many datasets require license purchase, and in many cases, these datasets are sold at remarkably high prices. Researchers find {that} it is difficult to secure funding to purchase their needed datasets. On the other hand, the free shared datasets are often presented in a researcher-unfriendly way. The developers of those datasets may either put too much trivial information or leave data unprocessed. Even worse, many datasets are shared online without providing basic descriptions. A lot of efforts have been made to improve the accessibility of our dataset. First, we create an easily accessible data webpage on GitHub where researchers can download our dataset and find the needed description and usage instructions. Second, there are also no specific restrictions on our data usage. Moreover, we hope our data can be easily used by researchers from different backgrounds, so it is presented in a researcher-friendly manner. For example, all non-visual contents are processed into tabular attributes and organized into separate tables according to their categories. All the car models have been assigned a unique identifier (ID), which can be used for information integration and future data fusion. In addition, we remove the background of collected images, which can simplify the potential applications for researchers interested in automotive exterior design.

\textbf{Quality} refers to missing data and error values. If problems exist in the original data, they can hardly be resolved by researchers using the data. Therefore, a series of data cleaning procedures are adopted to ensure the resulting data quality. First, the ambiguous or inconsistent values are unified or corrected. For example, a car's brand or name containing \lq\lq{}Benz\rq\rq{} is corrected into \lq\lq{}Mercedes\rq\rq{}. Second, data values at different granularities are aggregated according to certain groups. For instance, annual sales regarding various car trims, a trim level representing the equipment levels in a specific car model, are aggregated to the model levels. Third, uncertain contents are largely abandoned. The raw collected data has more than six million images, but most of them are abandoned in later processes for the purpose of quality control.

Besides the aspects above, ethics issues are thoughtfully reviewed throughout the data preparation. We carefully pre-process and reproduce image contents to strictly comply with the GDPR (e.g., removing image background and covering plate numbers). It is worth noting that the UK government has special laws to encourage research studies related to creating and using web content based datasets\footnote{\href{https://www.gov.uk/guidance/exceptions-to-copyright\#non-commercial-research-and-private-study}{https://www.gov.uk/guidance/exceptions-to-copyright}}. Therefore, we do not have copyright concerns of our collected data if people use it for non-commercial purposes. As our dataset only contains car-related information (e.g., images, car model specification, sales), it seems unlikely it will generate negative societal impact.

\section{The DVM-CAR Dataset}
\label{sec:dvm_car}

\begin{figure*}[htp]
\centering
\begin{subfigure}[b]{0.32\linewidth}
\centering
\includegraphics[width=1\linewidth]{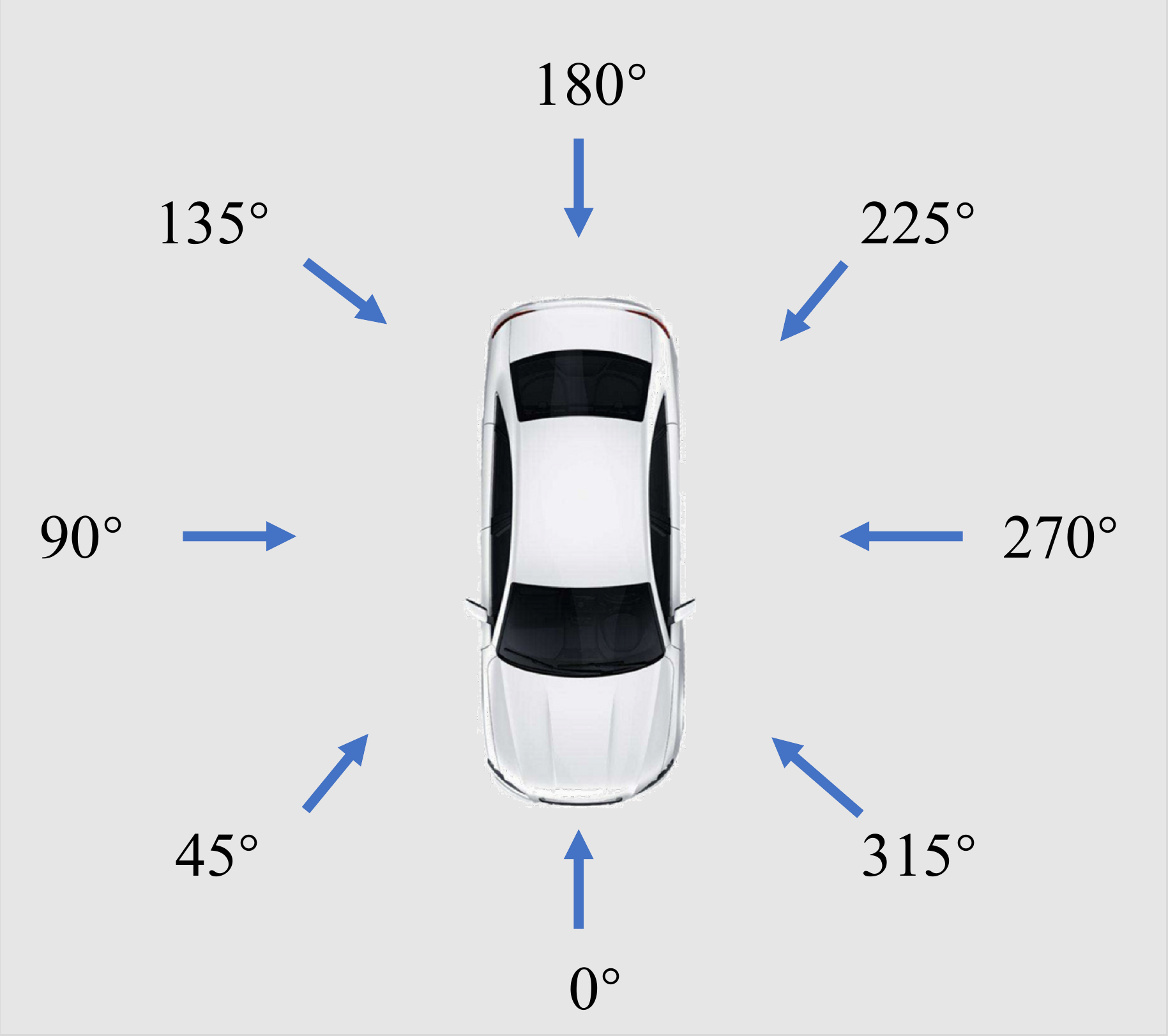}
\caption{}
\end{subfigure}
\hfill
\begin{subfigure}[b]{0.32\linewidth}
\centering
\includegraphics[width=1\linewidth]{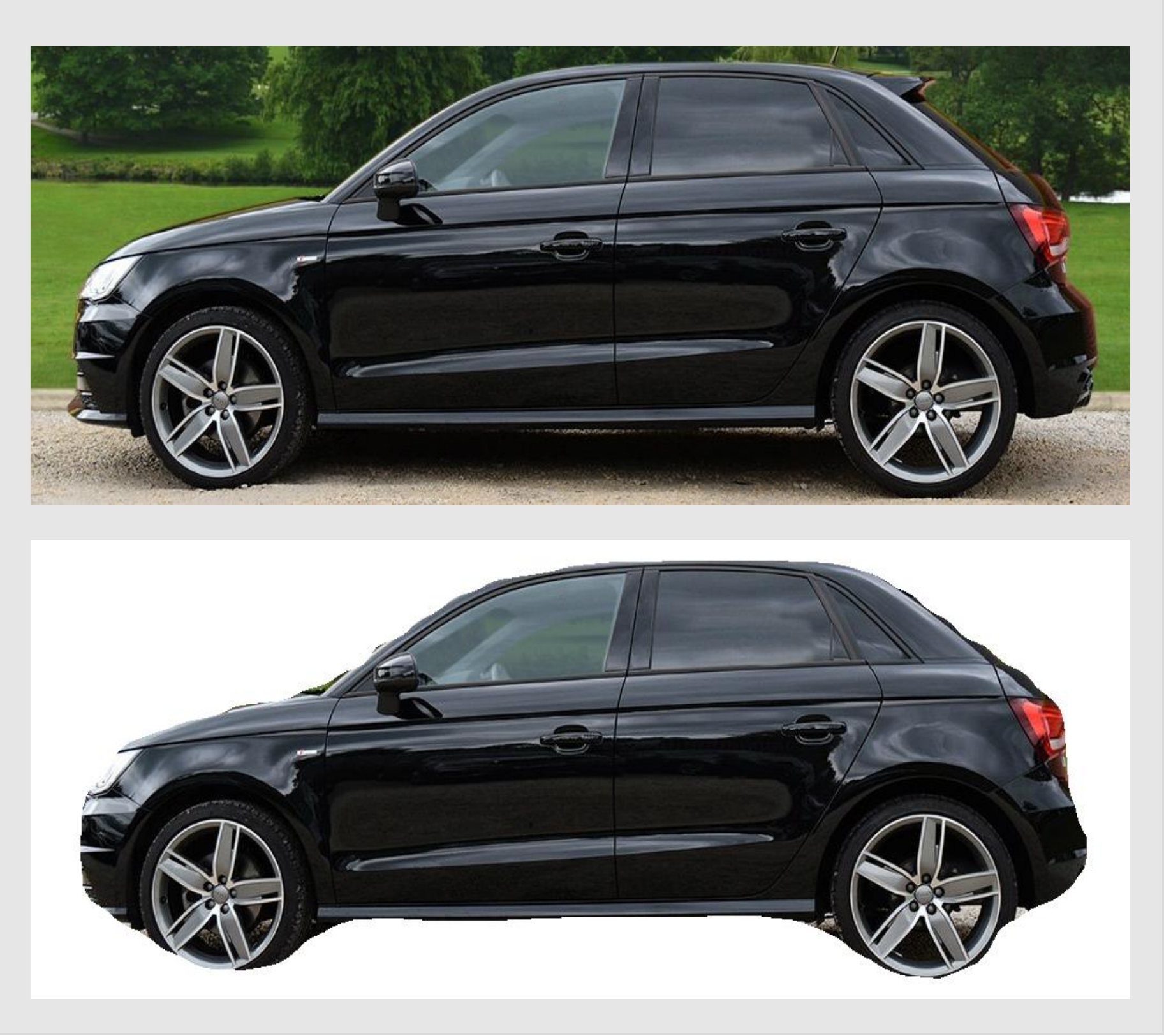}
\caption{}
\end{subfigure}
\hfill
\begin{subfigure}[b]{0.32\linewidth}
\centering
\includegraphics[width=1\linewidth]{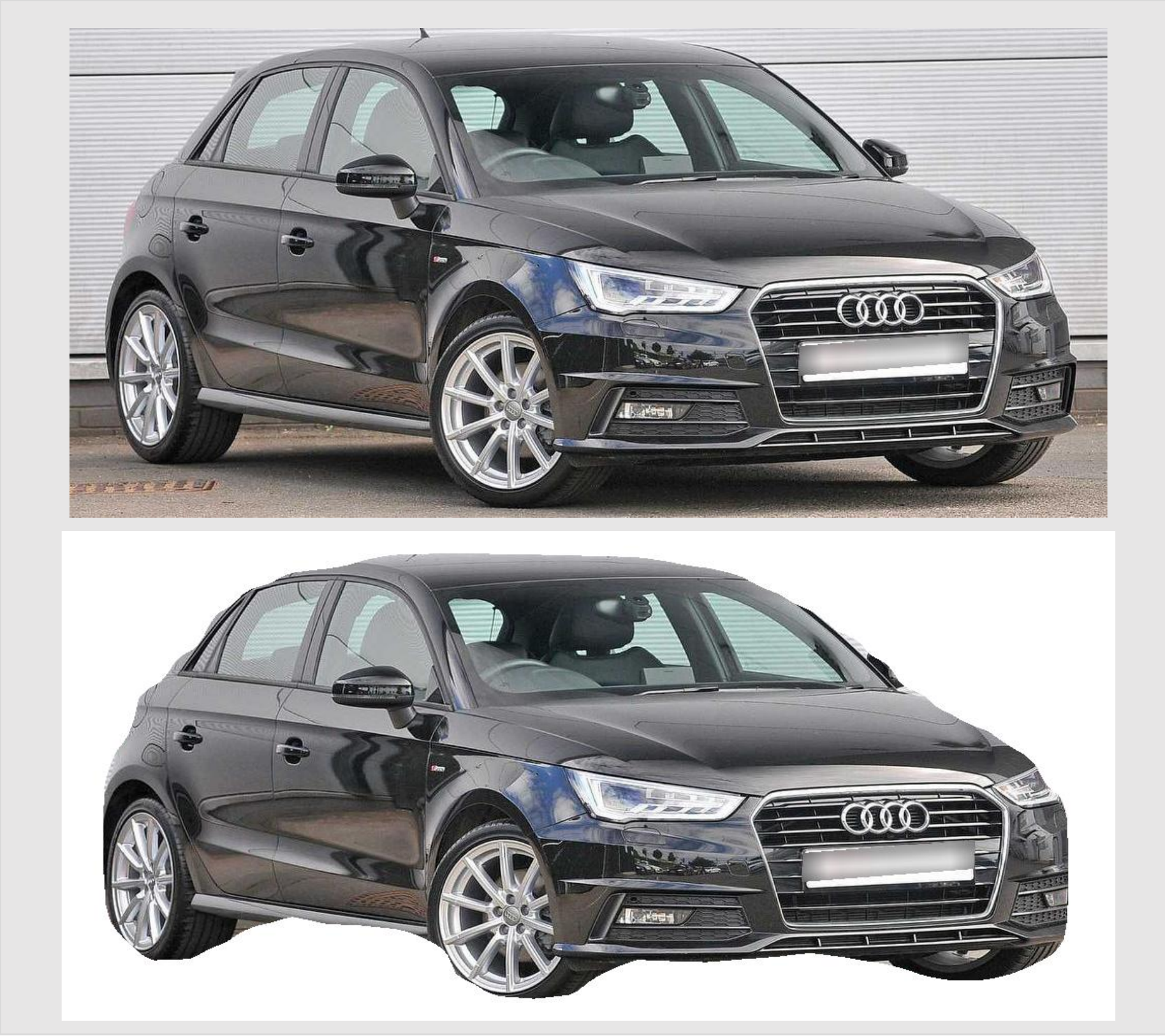}
\caption{}
\end{subfigure}
\caption{Illustration of car images in our dataset: (a) angles of car image labeling; (b) example of background {removal} for Audi A2 from angle 270 degree; (c) example of background {removal} for Audi A2 from angle 45 degree.}
\label{fig:three_graphs}
\end{figure*}
\begin{figure*}
\begin{subfigure}[b]{0.49\linewidth}
\centering
\includegraphics[width=\linewidth]{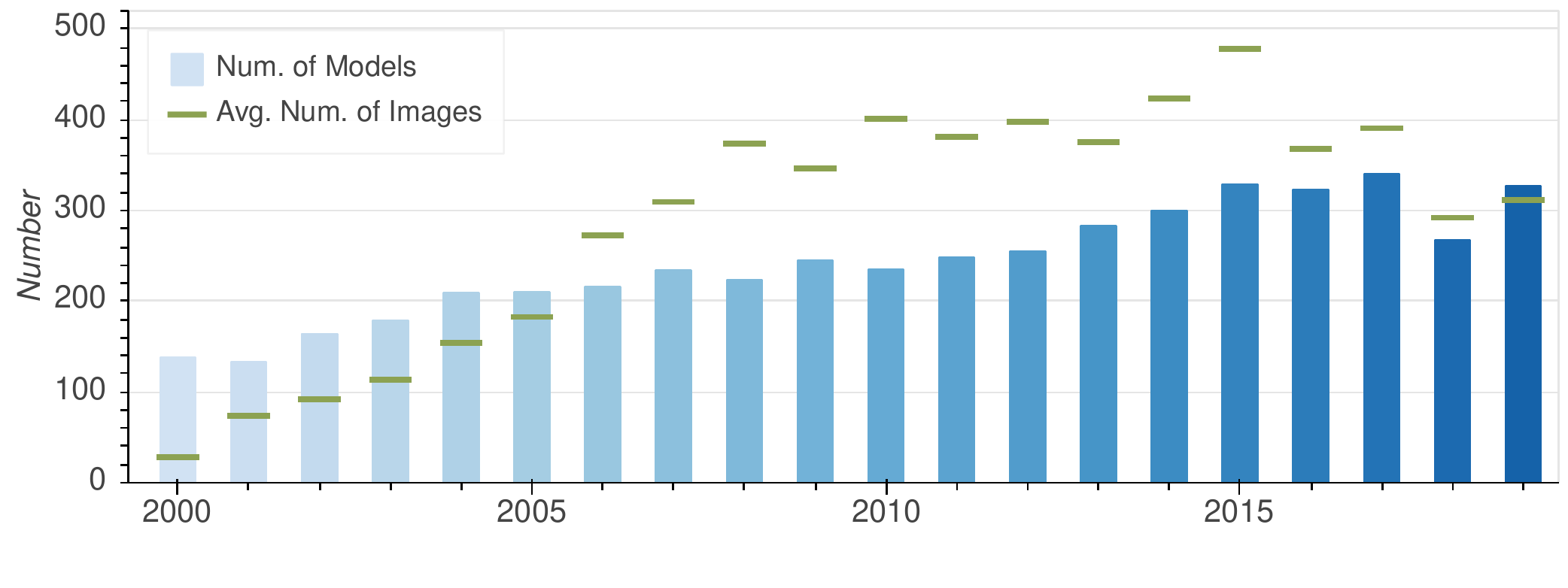}
\caption{}
\end{subfigure}
\begin{subfigure}[b]{0.50\linewidth}
\centering
\includegraphics[width=\linewidth]{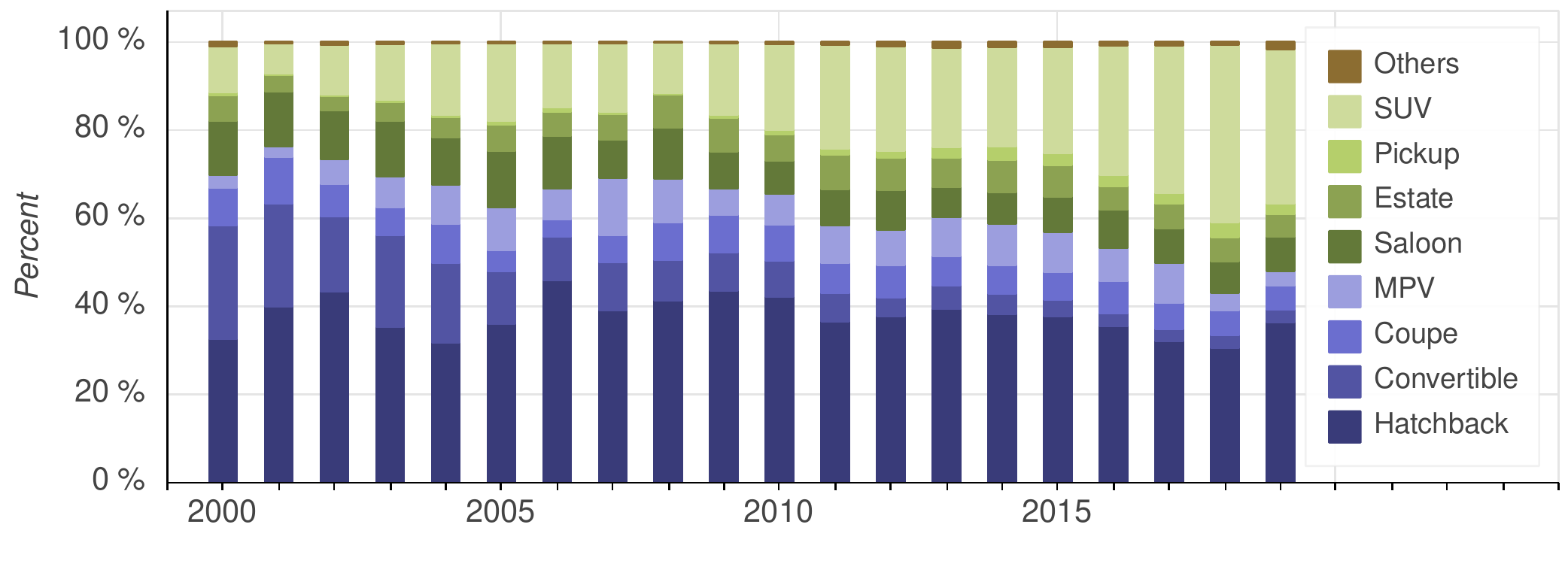}
\caption{}
\end{subfigure}
\caption{Summary of image data in the DVM-CAR dataset: (a) the number of car models across years and the average images for each model; (b) the annual percentage composition of images by body type. }
\label{fig:image_stat_grid}
\end{figure*}

As our primary motivation to facilitate the visual marketing research and forecasting applications, we call the proposed dataset \textbf{Deep Visual Marketing Car} (in short \textbf{DVM-CAR}), which is publicly available under the \textbf{CC BY-NC} license at:

\vspace{0.3em}
\begin{minipage}{2cm}
\setlength{\parindent}{2.5em}
\href{https://deepvisualmarketing.github.io}{https://deepvisualmarketing.github.io}. 
\end{minipage}
\vspace{0.3em}

The DVM-CAR dataset consists of two parts: image data and table data. The part of \textbf{image data} contains 1,451,784 car images (in JPEG format with resolution 300$\times$300) which are compressed in a ZIP file (13.6 GB file size). These images have been selected from eight observation viewpoints and stored under the categorization \lq\lq{}brand-model-year-color\rq\rq{}. This structure allows researchers to locate target images easily. The backgrounds of all car images are removed. A table is prepared for image indexing, so researchers can select the images via this table. In total, our image data covers 899 car models that sold in the UK market over the last 20 years. As Fig.~\ref{fig:image_stat_grid} (a) shows, the automotive classified advertising platforms have more data for newly-launched models than older car models. In spite of this, it contains 138 car models sold in the year 2000, each with an average of 28 images. The wide longitude of the dataset makes the observation of long-term trends easy, thus, is particularly useful for economic and marketing analytics and forecasting. For instance, through Fig.~\ref{fig:image_stat_grid} (b), it is shown that SUVs are becoming more popular and taking higher market shares over the last two decades. The part of \textbf{table data} (i.e., non-visual part) comprises six tables in the CSV format (156 MB file size), which are named the basic, sales, price, trim, ad and image tables. Together, these tables cover various variables and form a relational database~\cite{Codd1983}, as each two of them can be joined via the primary or secondary keys.More details of these tables can be found in Table~\ref{tab:table_desc}.  


\begin{table*}[htp]
\small
\centering
\caption{Description of table data in the DVM-CAR dataset.}
\label{tab:table_desc}
\adjustbox{max width=1\textwidth}{%
\begin{tabular}{l|l|l|l}
\toprule
Name                   & Table description                                                                                                                                                                                                                                                                                                                           & Attribute name      & Attribute description\\
\toprule
\multirow{4}{*}{Basic} & \multirow{4}{*}{\begin{tabular}[c]{@{}l@{}}It is mainly for indexing other  tables.\end{tabular}}                                                                                                                                                                                                                                         & Genmodel            & Generic model name                       \\ \cline{3-4} 
                             &                                                                                                                                                                                                                                                                                                                                             & Genmodel ID         & Generic model ID                                         \\ \cline{3-4} 
                             &                                                                                                                                                                                                                                                                                                                                             & Automaker           & Automaker name                             \\ \cline{3-4} 
                             &                                                                                                                                                                                                                                                                                                                                             & Automaker ID        & Automaker ID                                \\ \hline
\multirow{16}{*}{Ad}         & \multirow{16}{*}{\begin{tabular}[c]{@{}l@{}}It contains more than 0.27 million used car advertisements\\
information  posted on the automotive classified advertising\\
platforms, including variables like advertisement's creation \\ time, 
used car registration year, cumulative mileage, selling  \\price, etc.\end{tabular}} & Genmodel            & Generic model name                               \\ \cline{3-4} 
                             &                                                                                                                                                                                                                                                                                                                                             & Genmodel ID         & Generic model ID                                  \\ \cline{3-4} 
                             &                                                                                                                                                                                                                                                                                                                                             & Maker               & Automaker name                                 \\ \cline{3-4} 
                             &                                                                                                                                                                                                                                                                                                                                             & Adv ID              & Advertisement ID                                                \\ \cline{3-4} 
                             &                                                                                                                                                                                                                                                                                                                                             & Adv year            & Advertisement's creation year                                      \\ \cline{3-4} 
                             &                                                                                                                                                                                                                                                                                                                                             & Adv month           & Advertisement's creation month                                     \\ \cline{3-4} 
                             &                                                                                                                                                                                                                                                                                                                                             & Color               & This car's color                               \\ \cline{3-4} 
                             &                                                                                                                                                                                                                                                                                                                                             & Reg year            & This car's first registration/selling year                         \\ \cline{3-4} 
                             &                                                                                                                                                                                                                                                                                                                                             & Bodytype            & This car's body type                                  \\ \cline{3-4} 
                             &                                                                                                                                                                                                                                                                                                                                             & Runned Miles        & This car's runned mileage                                    \\ \cline{3-4} 
                             &                                                                                                                                                                                                                                                                                                                                             & Engin size          & This car's engin size                                \\ \cline{3-4} 
                             &                                                                                                                                                                                                                                                                                                                                             & Gearbox             & This car's gearbox                                    \\ \cline{3-4} 
                             &                                                                                                                                                                                                                                                                                                                                             & Fuel type           & This car's fuel type                             \\ \cline{3-4} 
                             &                                                                                                                                                                                                                                                                                                                                             & Price               & This car's selling price 
                             \\ \cline{3-4} 
                             &                                                                                                                                                                                                                                                                                                                                             & Seat num            & This car's seats number                           \\ \cline{3-4} 
                             &                                                                                                                                                                                                                                                                                                                                             & Door num            & This car's doors number                           \\ \hline
\multirow{5}{*}{Image} & \multirow{5}{*}{\begin{tabular}[c]{@{}l@{}}It contains image data related information like predicted \\viewpoint  and quality check result.\end{tabular}}                                                                                                                                                                                 & Genmodel ID         & Generic model ID                               \\ \cline{3-4} 
                             &                                                                                                                                                                                                                                                                                                                                             & Image ID            & Image ID                                       \\ \cline{3-4} 
                             &                                                                                                                                                                                                                                                                                                                                             & Image name          & Image name                                        \\ \cline{3-4} 
                             &                                                                                                                                                                                                                                                                                                                                             & Predicted viewpoint & This image's predicted viewpoint                   \\ \cline{3-4} 
                             &                                                                                                                                                                                                                                                                                                                                             & Quality check       & Manually check result                                        \\ \hline
\multirow{5}{*}{Price}       & \multirow{5}{*}{\begin{tabular}[c]{@{}l@{}}It contains the entry-level new car prices. It is designed for  \\ people who only need the basic price of  car models.\end{tabular}}                                                                                                                                                         & Genmodel            & Generic model name                                       \\ \cline{3-4} 
                             &                                                                                                                                                                                                                                                                                                                                             & Genmodel ID         & Generic model ID                                         \\ \cline{3-4} 
                             &                                                                                                                                                                                                                                                                                                                                             & Maker               & Automaker name                                  \\ \cline{3-4} 
                             &                                                                                                                                                                                                                                                                                                                                             & Year                & Generic model's selling year                                                   \\ \cline{3-4} 
                             &                                                                                                                                                                                                                                                                                                                                             & Entry price         & Generic model's entry-level price \\ \hline
\multirow{4}{*}{Sales}       & \multirow{4}{*}{\begin{tabular}[c]{@{}l@{}}It contains car sales data of the UK market (based on the \\ released statics from the DVLA).\end{tabular}}                                                                                                                                                                                    & Genmodel            & Generic model name                                               \\ \cline{3-4} 
                             &                                                                                                                                                                                                                                                                                                                                             & Genmodel ID         & Generic model ID                                              \\ \cline{3-4} 
                             &                                                                                                                                                                                                                                                                                                                                             & Maker               & Automaker name                        \\ \cline{3-4} 
                             &                                                                                                                                                                                                                                                                                                                                             & Year 2001 to 2020   & Generic model's annual sales                            \\ \hline
\multirow{9}{*}{Trim}        & \multirow{9}{*}{\begin{tabular}[c]{@{}l@{}}It includes 0.33 million trim level information such as  sell\\-ing sellingprice, fuel type and engine size.  It is designed   \\for people who  are interested in the price of car model at\\   a specific trim level.\end{tabular}}                                                                & Genmodel            & Generic model name                                      \\ \cline{3-4} 
                             &                                                                                                                                                                                                                                                                                                                                             & Genmodel ID         & Generic model ID                                          \\ \cline{3-4} 
                             &                                                                                                                                                                                                                                                                                                                                             & Trim                & Trim name                                             \\ \cline{3-4} 
                             &                                                                                                                                                                                                                                                                                                                                             & Maker               & Automaker name                                        \\ \cline{3-4} 
                             &                                                                                                                                                                                                                                                                                                                                             & Year                & Trim's selling year                                     \\ \cline{3-4} 
                             &                                                                                                                                                                                                                                                                                                                                             & Price               & Trim's price at selling year\\ \cline{3-4} 
                             &                                                                                                                                                                                                                                                                                                                                             & Gas emission        & Trim's CO$_2$ emission                                  \\ \cline{3-4} 
                             &                                                                                                                                                                                                                                                                                                                                             & Fuel type           & Trim's fuel type                                  \\ \cline{3-4} 
                             &                                                                                                                                                                                                                                                                                                                                             & Engine size         & Trim's engine size                                     \\ 
\bottomrule
\end{tabular}
}
\end{table*}



\section{Application Examples} 

This section briefly illustrates three application examples for demonstrating how the DVM-CAR dataset could be applied to business research and applications.

\subsection{Understanding Automotive Exterior Aesthetics Design} 

Product aesthetics design is a determinant of consumer acceptance and product success~\cite{hoffer1984automobile,bloch1995seeking,Schoormans1997,jindal2016designed}. Marketing scholars have discussed aesthetics from various aspects, including the influence of aesthetics on product differentiation and new product development, and specific determinants of consumer responses to aesthetics. For example, morphing techniques were used to quantify and incorporate aesthetics design into empirical car sales models~\cite{landwehr2011, Tseng2018}. This is an important step in modeling the effect of aesthetics design on sales. However, the used method is still limited in its ability to process image data. Car visual attributes are only extracted by pre-defined feature extractors, which are coarse-grained, and their respective implications for sales analysis and car appearance design are limited.

The recent advancement of machine learning technologies has provided marketing researchers with new tools to investigate product aesthetics. Several studies applied deep neural networks such CNNs to interpret the perceived design features~\cite{pan2016quantitative, burnap2016improving}. Compared with the traditional quantitative methods used in marketing research, deep learning algorithms can automatically learn high-level representations of visual features from car image data. Thus, the deep models can be used end-to-end, which solely requires raw images and tagging data as labels. Such advantages of deep models make them widely applied for visual based predictive studies~\cite{lecun2015deep}.

The DVM-CAR dataset provides an excellent base for researchers to apply deep learning to extract visual attributes from car images. For instance, Fig.~\ref{fig:fig_application_samples} (b) shows the sample using car images to infer the modernity score (i.e., labels computed from car registration years). By fine-tuning the ImageNet pre-trained VGG'16~\cite{Simonyan15}, the results show deep learning models are capable of predicting the design fashion for family cars. Besides, we can further recognize the design language or patterns from the inside gradients. Fig.~\ref{fig:fig_application_samples} (a) displays the predicted scores for Land Rovers. Although all these models are from the same automaker and were sold in the same year, the trained deep model rates them with entirely different modernity scores. The second row of Fig.~\ref{fig:fig_application_samples} (a) presents, with the help of visualization methods such as guided back-propagation~\cite{springenberg2014striving}, researchers can identify specific designs (i.e., highlighted in blue color) that make the appearance outdated.

Based on the car models' modernity predictions, we can further investigate the associations between cars' modernity of appearance and their market performance by using the sales records in the DVM-CAR dataset. As Table~\ref{tab:survival_chances} illustrates, we find the models' future withdrawn chances are correlated to their modernity scores (Note: the numbers in brackets of Table~\ref{tab:survival_chances} are the sample sizes of the groups). The car group with low modernity scores have a higher chance of being withdrawn from the market, while the high modernity group tend to survive longer. As essential marketing variables such as new car prices and sales are all covered by the dataset, users can investigate the association between cars' exterior styling and market performance from diverse perspectives.

\begin{figure*}[hp]
\centering
\includegraphics[width=1\linewidth, trim={4cm 19cm 2.5cm 4.5cm},clip]{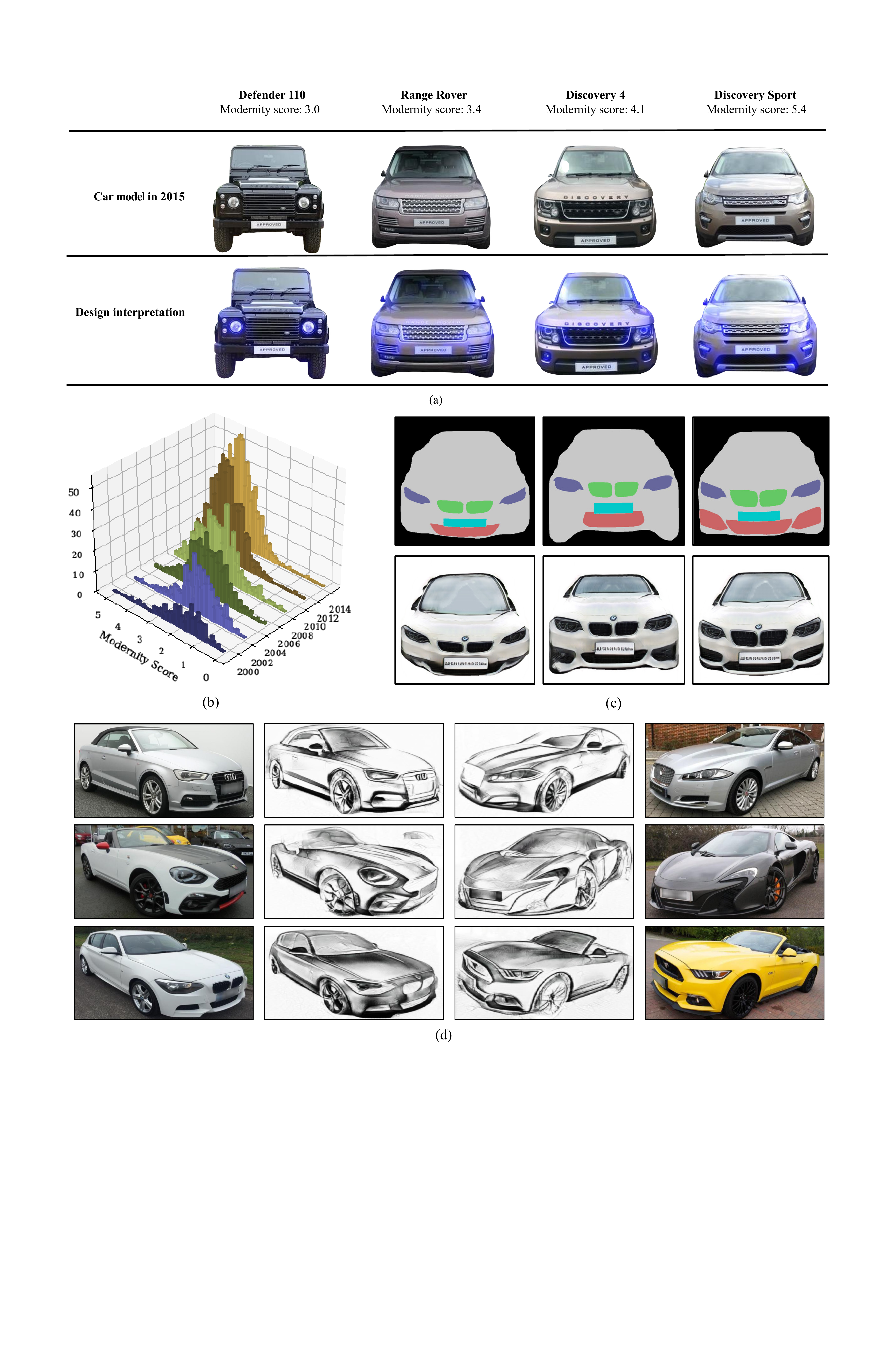}
\caption{Illustration of application examples: (a) modernity scores of Land Rover models registered in 2015; (b) distribution of the predicted modernity scores of car models from 2000 to 2015; (c) new BMW front designs generated by CycleGAN according to given facial shapes; (d) automobile design sketches generated by CycleGAN.}
\label{fig:fig_application_samples}
\end{figure*}

\subsection{AI-Powered Automotive Exterior Design}

Anthropomorphism~\cite{miesler2011isn,waytz2014mind,Ku2014} refers to the attribution of human or animal characteristics to non-living objectives. As a common phenomenon in consumption environments, anthropomorphism has drawn wide attention from both social psychology and marketing researchers. Psychological studies~\cite{purucker2014consumer} reveal that our brain is highly specified for face perception, which is so evolved that we often perceive faces from non-living objects. A group of marketing studies extensively investigate how products with human-like or animal characteristics lead to face perception. Typically, in car related studies, existing investigations~\cite{aggarwal2007car,Landwehr2011a, miesler2011isn} show consumers have a strong tendency to anthropomorphize the car front. However, in these studies, human or animal characteristics are predefined and only used in user surveys. They cannot be further deployed for facial feature recognition for unseen images. Thus, providing limited insights on car front appearance design for car manufacturers. 


\begin{figure*}[t]
\centering 
\includegraphics[width=0.95\linewidth, trim={1.9cm 8.55cm 2.05cm 4.45cm},clip]{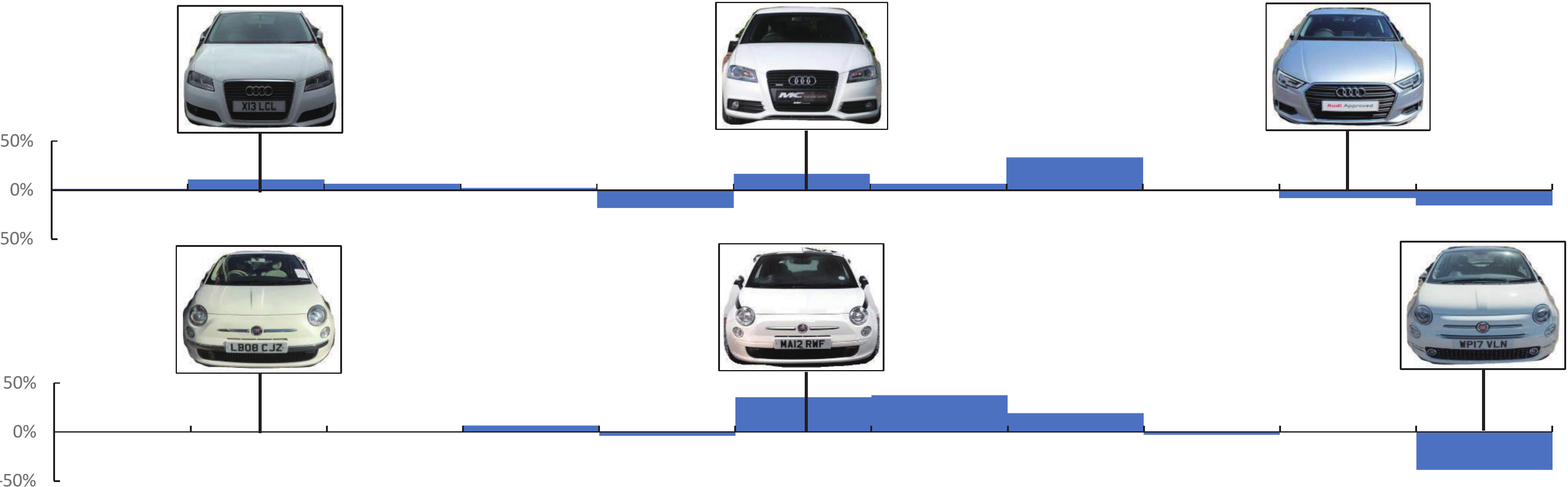}
\caption{Predicting the changes of market share changes according to design changes. The displayed two samples are the historical Audi A3 and Fiat 500 in the last decade. The position of displayed car images indicates the year for model redesign, the blue bar indicates the predicted market share change.}
\label{fig:sales_changes}
\end{figure*}

As a group of generative models with neural network structures, deep generative models can learn to generate highly realistic representations from data, thus making them a hotspot in machine learning and statistics in past years~\cite{goodfellow2014generative}. These models have become extremely successful in the applications such as image translation and fake data generation. Particularly, marketing researchers apply these methods to generate novel product designs~\cite{Arbor2016, quan2018product, Yu2019}, including cars. Pan et al.\cite{pan2017deep} and Burnap et al.\cite{Burnap2019} used the deep generative models in the combination of design evaluators (i.e., deep convolution models trained for aesthetics evaluating) to generate appealing car appearance designs. The 1.4 million car images of the DVM-CAR dataset make it extremely useful for generative model training. Two examples are provided in Fig.~\ref{fig:fig_application_samples} (c) and (d). The former shows new BMW 3 Series front designs generated by cycle-consistent adversarial networks (in short CycleGAN)~\cite{zhu2017unpaired} based on the fed semantic views. The latter presents sketches generated from real car images via the CycleGAN. These are from our working topics where we try to obtain bio-inspired designs by morphing the cars' layouts after cheetah faces.

\subsection{Visual-based Car Sales Forecasting}

Several existing studies~\cite{Green1981,Science1987,Kaul1995,Shi2001} investigate the product exterior design as an optimization problem that searches for best-selling designs with considering aesthetic attributes as controllable factors in manufacturing. For example, Rajeev and Krishnamurti~\cite{Science1987}  simulated market sales using the share-of-choice setting that viewed each product as a profile with a collection of design attributes at different discrete levels. Due to their heterogeneous needs, the consumers are willing to pay more for the product with matched profile and thus more likely to buy. Differently, Kaul and Rao~\cite{Kaul1995} considered consumers' heterogeneous needs as coordinates distributed in a high-dimensional space of product attributes, where utilities of products are indicated by their distances from the coordinates. Their study tried to infer the best coordinates in the market that maximize the overall purchase likelihood. However, as designs are represented by the profiles in these existing studies, no actual designs are generated. The complexity of real scenarios is thus too simplified to be applicable. For instance, the existing studies didn't clarify how to compute an aesthetic profile of a given design and whether particular aesthetic profiles are not achievable (e.g., a super sporty and cute design).

One of our working studies incorporates the deep learning-based design evaluation and generation procedure into product design optimization. Historical images and sales of the DVM-CAR are used to predict the changes in sales when adopting a new design. Being trained with massive car images, the GAN model is capable of generating various novel yet highly realistic car designs. Meanwhile, after collecting aesthetic ratings from surveys, a CNN model is trained with these aesthetic scores and car images to assess designs in terms of aesthetics. Then the trained CNN is used to evaluate both real designs from the dataset and the ones generated by the GAN model. For sales forecasting, an RNN model is trained with these obtained aesthetic scores and annual sales records. Thus, recognizing a car model's prior sales and its aesthetic changes, the trained RNN can forecast the posterior sales (or market shares)., see Fig.~\ref{fig:sales_changes} for a series of the predicted market share changes of Audi A3 and Fiat 500.

\begin{table}[t]
\small
\centering
\caption{Withdrawn chances within the next three years of car groups with different modernity scores.}
\label{tab:survival_chances}
\adjustbox{max width=1\linewidth}{%
\begin{tabular}{l|c|c|c|c|c}
\toprule
\multirow{2}{*}{Year} & \multicolumn{5}{c}{Modernity score}\\
\cline{2-6}
       & {[}0,1)    & {[}1,2)                     & {[}2,3)    & {[}3,4)   & {[}4,5{]} \\
\toprule
2000 & 12.0\%(25) & 2.8\% (36)                  & 0.0\% (10) & -(2)      & -(0)      \\

2003 & 23.1\%(13) & 10.7\%(56) & 0.0\%(28)  & 0.0\%(10)      & -(0)      \\

2006 & -(4)  & 13.9\%(36) & 8.3\%(60)  & 7.4\%(27) & -(1)      \\

2009 & -(1)       & 33.3\%(9)                  & 16.1\%(62) & 3.3\%(61) & 15.8\%(19) \\

2012 & -(0) & -(1) & 17.9\%(28) & 9.1\%(88) & 5.4\%(56) \\

2015 & -(0) & -(0) & 18.2\%(11) & 9.0\%(67) & 7.3\%(109)\\
\bottomrule
\end{tabular}
}
\end{table}

\section{Conclusion}
\label{sec:conclusion}

This paper demonstrates the design and development of a large-scale dataset for business research with online sources. As highlighted in Section~\ref{sec:intro}, our work has significant contributions for information and knowledge management, ranging from domain application to big data, to database design, and to data fusion. The targeted users of our dataset are business researchers and computer scientists who work in visual-related research and applications, particularly on (but not limited to) the topics of automotive exterior design, consumer analytics and sales prediction. Through this paper, we would like to make an announcement that the DVM-CAR is now released publicly to research communities. We hope it can have a large impact. More importantly, we would like to maintain and keep updating the dataset by incorporating researcher's feedback over time. 

\section*{Acknowledgment}
The first author acknowledges the Adam Smith Business School and the {School of} Computing Science of University of Glasgow's funding support for this research. The second author would like to thank the funding support of the Region Bourgogne Franche Comt\'e Mobility Grant, Nvidia Accelerated Data Science Grant and Google Cloud Academic Research Grant.

\bibliographystyle{IEEEtran}
\bibliography{IEEEabrv,mybib}
\end{document}